\documentclass[10pt,reqno]{amsart} 

\usepackage{general,stmaryrd}
\usepackage[activate={true,nocompatibility},final]{microtype}
\usepackage[left=1.5in,right=1.5in,top=1.2in,bottom=1in]{geometry}%showframe
\usepackage[bookmarks, colorlinks=true, pdfstartview=FitV, linkcolor=blue, citecolor=blue, urlcolor=blue]{hyperref} %
\usepackage{cite}
\usepackage[dayofweek]{datetime}
	\settimeformat{ampmtime} \usdate

  \let\oldmarginpar\marginpar
  \renewcommand\marginpar[1]
  {\-\oldmarginpar[
    \begin{minipage}{\marginparwidth}\raggedleft\tiny #1 \end{minipage}] 
   {\begin{minipage}{\marginparwidth}\raggedright\tiny #1 \end{minipage}}
  }

\newcommand{\ds}{\operatorname{d}\negsp[3]s}
\newcommand{\dt}{\operatorname{d}\negsp[3]t}

\renewcommand{\ve}[1][x]{\ensuremath{\mathbf{#1}}\xspace}

\renewcommand{\exp}{\textrm{exp}}

\newcommand{\Nystrom}{Nystr\"om\xspace}
\newcommand{\known}{\ensuremath{A}\xspace}

\begin{document}

	\title[Iterated geometric harmonics]{Iterated geometric harmonics for data imputation and reconstruction of missing data}

  \author[Eckman, Lindgren, Pearse, Sacco, Zhang]{Chad~Eckman, Jonathan~A.~Lindgren, Erin~P.~J.~Pearse, David~J.~Sacco, Zachariah~Zhang}
  
  \thanks{The authors are grateful to support from the College of Science and Mathematics Summer Research Program
at Cal Poly, San Luis Obispo. %All authors are with the Department of Mathematics, California Polytechnic State University, San Luis Obispo, CA 93407-0403.	
	}

%  \markboth{IEEE Transactions on Pattern Analysis and Machine Intelligence,~Vol.~?, No.~?, ?~2014}
%{Eckman \MakeLowercase{\textit{et al.}}: Iterated geometric harmonics and incomplete data}

%\IEEEpubid{Version of \textbf{\currenttime} on  \textbf{\longdate{\today}}. \hstr[3] --- \hstr[3] 0000--0000/00\$00.00÷\copyright÷2014 IEEE}

  %\author{Chad Eckman}
  %\address{California Polytechnic State University, Mathematics Department, San Luis Obispo, CA 93407-0403 USA}
  %\email{\href{mailto:ceckman@calpoly.edu}{ceckman@calpoly.edu}}

  %\author{Jonathan Lindgren}
  %\address{California Polytechnic State University, Mathematics Department, San Luis Obispo, CA 93407-0403 USA}
  %\email{\href{mailto:jlindgre@calpoly.edu}{jlindgre@calpoly.edu}}

  %\author{Erin P. J. Pearse}
  %\address{California Polytechnic State University, Mathematics Department, San Luis Obispo, CA 93407-0403 USA}
  %\email{\href{mailto:epearse@calpoly.edu}{epearse@calpoly.edu}}

  %\author{David J. Sacco}
  %\address{California Polytechnic State University, Mathematics Department, San Luis Obispo, CA 93407-0403 USA}
  %\email{\href{mailto:djsacco@calpoly.edu}{djsacco@calpoly.edu}}

  %\author{Zachariah Zhang}
  %\address{California Polytechnic State University, Mathematics Department, San Luis Obispo, CA 93407-0403 USA}
  %\email{\href{mailto:zazhang@calpoly.edu}{zazhang@calpoly.edu}}

%\IEEEtitleabstractindextext{%
\begin{abstract}
	The method of geometric harmonics is adapted to the situation of incomplete data by means of the iterated geometric harmonics (IGH) scheme. The method is tested on natural and synthetic data sets with 50--500 data points and dimensionality of 400--10,000. Experiments suggest that the algorithm converges to a near optimal solution within 4--6 iterations, at runtimes of less than 30 minutes on a medium-grade desktop computer. The imputation of missing data values is applied to collections of damaged images (suffering from data annihilation rates of up to 70\%) which are reconstructed with a surprising degree of accuracy.
\end{abstract}

\maketitle

%  \keywords{Geometric harmonics, incomplete data, missing data, imputation, out-of-sample extension of an empirical function, diffusion maps.}
%  \subjclass[2010]{
%    Primary: 
%    15-04, %Linear algebra: Explicit machine computation and programs
%    15A18, %Eigenvalues, singular values, and eigenvectors
%		15A83, %	Matrix completion problems
%    42-04, %Harmonic analysis: Explicit machine computation and programs
%    62G05. %Statistics: Estimation
%    Secondary: 
%    44-04, %Integral transforms: Explicit machine computation and programs
%    68T10, %Pattern recognition, speech recognition
%    68T30, %	Knowledge representation
%    65D15. %Algorithms for functional approximation
%    }
%  \thanks{}

\section{Introduction}

The method of geometric harmonics was introduced in the thesis of S. Lafon \cite{Lafon04} (see also \cite{GeometricHarmonics}) as a method of extending empirical functions defined on a dataset, considered as a point cloud $X = \{\ve[x]_i\}_{i=1}^n \ci \bRd$, where $d$ is the dimension (number of parameters or characteristics) observed for each data point (observation) $\ve[x]_i$. In other words, consider a subset $\known \ci \{1,\dots,n\}$ and let $X_\known = \{\ve[x]_i \in X \suth i \in \known\}$. Then given a function $f:X_\known \to \bR$, the goal is to construct an extension $\widehat{f}:X \to \bR$ for forecasting, classification, or other machine learning or statistical analysis purposes. 

In the present paper, we do not consider inferring the value of some extrinsic function defined on the data (as in a classification or clustering problem) but turn the mechanism on the dataset itself. 
I.e., if one considers $f$ as one column of the dataset, the task of extending the partially defined function $f:X_\known \to \bR$ amounts to imputing the missing values in a dataset where missing values occur only in that one column. In this paper, we introduce the method of iterated geometric harmonics (IGH) which uses an iteration scheme to adapt the method of geometric harmonics to more general situations of incomplete data (i.e., missing values in many/all columns). The method allows for reconstruction of datasets which are incomplete due to the presence of missing values (due to recording error, transmission loss, etc.). The datasets should have the property that they can be thought of as samples from some underlying topological manifold; i.e., the data should be comprised of one or more clusters of points which are ``close'' in the general sense of manifold learning theory. We focus on \emph{missing} data, not \emph{noisy} data; i.e., we assume that it is known which data need to be reconstructed.

Geometric harmonics is parameter-free in the sense that no regression model is assumed: the technique constructs the extension directly from the geometry of the dataset. Nonetheless, the analyst still must choose some way to measure similarities between data points (observations) and this amounts to determining a particular kernel for an integral operator, i.e., a positive semidefinite symmetric function $k:\bRd \times \bRd \to \bR_+$. Some care must be taken when choosing the kernel; we give some suggestions for kernel selection below. We test our approach  on natural and synthetic data sets and some conditions for convergence of the iteration are discussed. In most cases we find the iteration converges remarkably fast; typically no more than about 5 iterates are required.

\subsection{Applications of IGH}

IGH was developed for imputing missing data prior to scientific/statistical analysis; see the example of weather data given below. Conventional statistical software do not cope well with missing data; typically the analyst is required to invoke some imputation procedure or discard incomplete data points. Discarding incomplete data is a poor choice, as it can easily bias the remaining data or leave the analyst with too little data for a proper analysis; see \cite{Enders, LittleRubin02}. The present state-of-the-art technique for dealing with missing data is Multiple Imputation, in which multiple complete versions of the data are simulated by filling in missing entries in a purely stochastic manner; the analyses of these different simulated versions are then averaged (pooled). Unfortunately, this is essentially a linear technique and does not perform well when the data has a strongly nonlinear structure. See \cite{Enders, LittleRubin02, vanBuuren} for details.

It will be clear from the examples below that additional applications of IGH include image processing, especially video reconstruction, as video media naturally contains large numbers of similar images (frames). Some samples of reconstructed video appear online at \\
\hstr[5]\scalebox{1}{\url{http://www.calpoly.edu/~epearse/video.html}.}\\
Based on the given examples, IGH clearly has potential applications for security, law enforcement, and the military, as well as reconstruction of archival footage and other tasks.

\section{Iterated geometric harmonics}

\subsection{Geometric harmonics}

The term ``geometric harmonics'' refers to an eigenbasis for an incomplete Gram matrix, constructed empirically by means of a variant of the \Nystrom method; cf. \cite{GeometricHarmonics}. The \Nystrom method is a quadrature rule that effectively allows one to approximate the solution to an integral equation by subsampling: the integral is replaced by a sum of function values (the function is evaluated on some sample points), each one of which is weighted according to the quadrature rule. 
In the classical case, the subsample may be selected manually for maximal effectiveness; in modern applications (esp. to machine learning problems, note that the \Nystrom formula has been shown to be equivalent to kernel PCA projection \cite{Scholkopf98nonlinearcomponent}), the subsample may be selected randomly as part of a Monte Carlo scheme. For examples related to spectral partitioning problems in image segmentation, see \cite{FowlkesBelongieChungMalik,FowlkesBelongieMalik}; for examples related to the extension of empirical eigenfunctions to data outside an original sample, see \cite{Belkin03laplacianeigenmaps, BelkinNiyogiSindhwani06, BengioPaiementVincentDelalleauLeRouxOuimet, Lafon04}; for examples related to the efficiency of kernel-based machine learning methods, see\cite{WilliamsSeeger01, DrineasMahoney2005, GittensMahoney2013}. 

In the context of geometric harmonics, one does not get to choose the subsample: it consists of those rows \known of the data set for which the partially defined function $f:X_\known \to \bR$ is defined. For a dataset with $n$ records, the method of geometric harmonics requires computing an $n \times n$ Gram matrix $K$ with entries 
\begin{align}\label{eqn:Gram-entries}
	K(i,j) = k(\ve[x]_i,\ve[x]_j), 
\end{align}
where $k:\bRd \times \bRd \to \bR_+$ is some symmetric, nonnegative, and typically positive semidefinite function. For example, for homogeneous data sampled uniformly from a submanifold of $\bRd$, we have used the Gaussian kernel
\begin{align*}%\label{eqn:}
	k(\ve[x],\ve[y]) &= \frac1{\gs\sqrt{2\gp}}\exp(-\tfrac12\|\ve[x]-\ve[y]\|_2^2/\gs^2).
\end{align*}
While the positive semidefinite property is not strictly necessary, it puts the method on firm theoretical footing within the context of reproducing kernel Hilbert spaces; cf. \cite{Aronszajn}. 

Geometric harmonics applies the \Nystrom method to the eigenequations
\begin{align}\label{eqn:eigenequations}
	\sum_{j} K(i,j) \gy(j) = \gl \gy(i).
	%\qq \ell=1,\dots,|\known|.
\end{align}
The eventual application of the eigenbasis is to the extension of a function $f:X_\known \to \bR$ to a function $\widehat{f}:X \to \bR$, in which case the operator $K:f \mapsto Kf$ is not a square matrix. Therefore, one must consider diagonalizing the $|\known| \times |\known|$ matrix $K^*K$, and it can be shown that the adjoint $K^*$ is simply the restriction operator; cf. \cite[Lem.~1]{GeometricHarmonics} or \cite[Lem.~13]{Lafon04}. This leads to solving \eqref{eqn:eigenequations} where $j$ ranges only over \known. However, it is observed in \cite[\S3.2]{Lafon04} that the left side of \eqref{eqn:eigenequations} makes sense for any $i \in \{1,\dots,n\}$; only $j$ needs to be restricted to \known. Thus, one can turn the equation around and \emph{define} the geometric harmonics $\gY_\ell$, for $\ell=1,\dots,|\known|$ by
\begin{align}\label{eqn:geometric-harmonics}
	\gY_\ell(i) := \frac1{\gl_\ell} \sum_{j \in \known} K(i,j) \gy_\ell(j).
\end{align}
The key point is that \eqref{eqn:geometric-harmonics} defines	$\gY_\ell(i)$ for \emph{all} $i=1,\dots,n$.
Formula \eqref{eqn:geometric-harmonics} generates a family of functions $\gY_\ell$ which are maximally concentrated on \known in a sense that generalizes the properties of the prolate spheroidal wave functions of Slepian; cf. \cite[\S2.2]{GeometricHarmonics} and \cite{MR0181766,MR0140732}.

The geometric harmonics now allow one to define the extension of $f:X_\known \to \bR$ by
\begin{align}\label{eqn:extension-algorithm}
	\widehat{f}(\ve[x]_i) := \sum_{\ell = 1}^{|\known|} \la f, \gy_\ell\ra_\known \gY_\ell(i),
\end{align}
where $\la f,g \ra_\known = \sum_{i \in \known} f(i) g(i) = f^T g$ is the restriction of the usual inner product to \known.
Note that \eqref{eqn:extension-algorithm} makes sense for \emph{any} row $\ve[x]_i$ of $X$, and in the case when $i \in \known$, formula \eqref{eqn:extension-algorithm} simply recovers the representation of the vector $f$ in terms of the ONB $(\gy_\ell)_{\ell=1}^{|\known|}$.

\subsection{The iteration scheme}
\label{sec:iteration-scheme}

Consider the \nth[j] coordinate function $\gg_j(\ve[x]_i) = \ve[x]_i^T \ve[e]_j = x_{ij}$ (where $\ve[e]_j$ is the standard unit basis vector with 1 in the \nth[j] entry and 0 elsewhere), and let $\known_j$ consist of those rows of $X$ for which $x_{ij}$ is \emph{not} a missing value. To address the problem of missing data, $\gg_j$ can be modeled as a function of the characteristics $m=1,\dots,j-1,j+1,\dots,p$. In this way, we can think of the \nth[j] column of $X$ as a partially-defined function $\gg_j:X_{\known_j} \to \bR$ to which the extension algorithm of geometric harmonics may be applied. %Note that this assumption may not be justified; it is easy to produce counterexamples by considering situations of extreme symmetry (think of a hypersphere centered at the origin). Nonetheless, t
This heuristic appears to be valid in practical examples, especially in applications to natural datasets with high dimensionality; cf. \S\ref{sec:results}.

The iteration scheme is initialized by stochastically imputing the missing values in the dataset. More precisely, for each $j$, the missing values of $\gg_j$ are drawn from a normal distribution with mean and variance computed equal to the sample mean and sample variance of $\gg_j |_{\known_j}$. For each iteration, the following steps are conducted for each $j=1$ to $d$:
\begin{enumerate}
	\item Form a new matrix $X^{(j)}$ by deleting column $j$ from $X$.
	\item Compute the (restricted) Gram matrix $K^{(j)}$ by applying \eqref{eqn:Gram-entries} to the rows of $X^{(j)}$.
	\item Define $\known_{j}$ to be the set of rows of $X$ for which $\gg_{j}$ is defined (i.e., for which $x_{ij}$ is \emph{not} a missing value).
	\item Compute the eigendata of ${K^{(j)}}^* K^{(j)}$, i.e., find $\gl_\ell^{(j)} \in \bR$ and $\gy_\ell^{(j)} \in \bR^{|\known_j|}$, that satisfy
	\begin{align*}%\label{eqn:geometric-eigendata}
		\sum_{m \in \known_j} K(i,m) \gy_\ell^{(j)}(m) = \gl_\ell^{(j)} \gy_\ell^{(j)}(i),
	\end{align*}
	for $\ell=1,\dots,|\known_j|$.
	Note that these functions $\gy_\ell^{(j)}$ are defined only for $i \in \known_j$.

	\item As in \eqref{eqn:geometric-harmonics}, construct the \emph{geometric harmonics} $\gY_\ell^{(j)}$ by 
	\begin{align*}%\label{eqn:}
		\gY_\ell^{(j)}(i) := \frac1{\gl_\ell^{(j)}} \sum_{m \in \known_{j}}  K(i,m) \gy_\ell^{(j)}(m).
	\end{align*}
	Note that these functions $\gY_\ell^{(j)}$ are defined for all $i$.

	\item As in \eqref{eqn:extension-algorithm}, fill in the missing values of $\gg_{j}$ using 
	\begin{align*}%\label{eqn:}
		\widehat{\gg}_j(i) 
		= \sum_{\ell=1}^{|\known_j|} \la \gg_{j},\gy_\ell^{(j)} \ra_{\known_{j}} \gY_\ell^{(j)}(i).
	\end{align*}
	%where 
	%\begin{align*}%\label{eqn:}
%		\la \ve[x]_{j_1},\gy_\ell\ra_{\known_{j}} := \sum_{x \in \known_{j}} \ve[x]_{j_1}(x) \gy_\ell(x). 
%	\end{align*}
	%Note here that the sum is over $x \in \known_{j}$, and so $\ve[x]_{j}(x)$ runs through those values of the first column which are known.
\end{enumerate}

\begin{remark}[Random shuffle of characteristics]\label{rem:shuffle}
	In an implementation of this method, the order in which the characteristics are considered is permuted after each iteration. In other words, let \gs be a random permutation of the characteristics $\{1,\dots,p\}$. Then, in each of steps (1)--(6), the index $j$ is replaced by the random index $\gs(j)$. This step prevents introducing a bias in the degree of correction corresponding to different indices. More precisely, we found during experimentation that introducing this step caused the algorithm to require 1--2 more iterations to achieve optimal convergence, but that final results were more accurate (in terms of the measurements discussed in section \S\ref{sec:results}). We omitted this detail in the description above, in an attempt to keep the notation from becoming overly heavy.
\end{remark}

\begin{remark}[Numerical difficulties]\label{def:Numerical-difficulties}
	It is clear from \eqref{eqn:geometric-harmonics} that the extension procedure can be ill-conditioned for $\gl_\ell \approx 0$; see \cite[\S3.2]{Lafon04}. This is taken into account in the implementation of the method.
\end{remark}

\section{Discussion and analysis of results}
\label{sec:results}

\subsection{Synthetic data}
\label{sec:synthetic-data}

To simulate data with intrinsic nonlinear geometry, we generated 250 points on a swiss roll in $\bR^3$. The points along the spirals were parameterized by arc length to ensure even spacing:
\begin{align*}%\label{eqn:}
	\ds = \tfrac12\left(\sqrt{1+t^2} + \sinh^{-1}(t)\right)\dt.
\end{align*}
We added height by generating 5 of these spirals on top one another, also evenly spaced. To enrich the dimensionality, the swiss roll was embedded into $\bR^{30}$, rotated in many random directions, and then Gaussian noise was added to ensure the dataset did not lie in any linear subspace $\bR^k$ with $k < 30$. The spread (rate of increase of distance to spiral), height, noise (variance), and number of rotations were chosen experimentally. This test dataset was designed to have such a uniform distribution of points so as to avoid any confounding influence of anisotropy while testing the method.

\begin{figure}%[h]
	$\negsp[50]$\begin{tabular}{ll}
	\scalebox{0.43}{\includegraphics{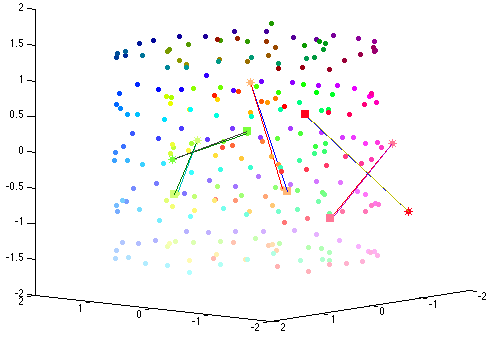}} 
		& \scalebox{0.65}{\includegraphics{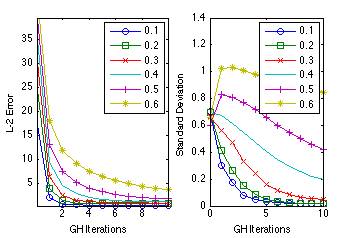}} \\
	\scalebox{0.43}{\includegraphics{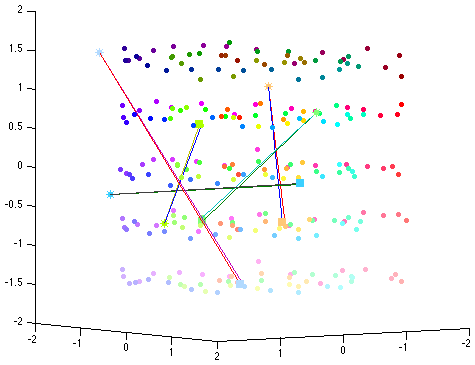}} &
	\scalebox{0.43}{\includegraphics{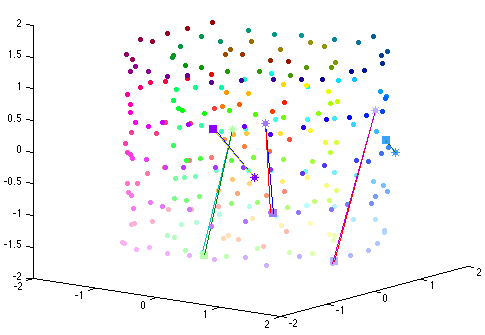}} \\
	\end{tabular}
	\caption{\captionsize The three scatter plots show the swiss roll dataset, reconstructed after annihilating data at rate $p=0.4$; see \S\ref{sec:synthetic-data}. The final imputed versions of each data point is plotted as a dot $\bullet$. However, for each plot, 5 data points $\ve[x]_i$, $i=1,\dots,5$, were selected for observation at random. The stochastically initialized point $\ve[x]_i^{(0)}$ is plotted as a $\star$, and the final imputed point $\ve[x]_i^{(end)}$ is plotted as a $\blacksquare$. The graphs at top right show the decay of $L^2$ error (left; lower values are better) and standard deviation (right) with each iteration; see \S\ref{sec:synthetic-data}.}
	\label{fig:swiss-roll-trails}
\end{figure}

Figure~\ref{fig:swiss-roll-trails} shows the results of some experiments with a synthetic dataset $X = (\ve[x]_i)_{i=1}^{250} \ci \bR^{30}$. The dataset $X$ was run through a program which deletes any value with fixed probability $p$. To assess the effectiveness visually, we used diffusion mapping $D_t:\bR^{30} \to \bR^3$ to embed the output of our experiments and produce 3-dimensional plots. For information on diffusion mapping (also introduced in Lafon's thesis; cf.~\cite[\S2]{Lafon04}), see the enjoyable introduction \cite{LafonLee06} and further references \cite{MR2238665, MR2724431, MR2447229, MR2443014, MR2238669}. We used a modified version of the excellent implementation of diffusion mapping provided by Laurens van der Maaten in his ``Matlab Toolbox for Dimensionality Reduction'' which can be found at \\
\phantom{m} \hfill \url{http://homepage.tudelft.nl/19j49/}. \hfill \phantom{m} \\
The imputed datasets in Figure~\ref{fig:swiss-roll-trails} visually match the original datasets quite well (although this is much easier to see while rotating the figures on the computer). 

Let $\ve[x]_i^{(k)}$ denote the imputed version of the \nth[i] datapoint $\ve[x]_i$ at iteration $k$ of the algorithm described in \S\ref{sec:iteration-scheme}. In order to track the effectiveness of the iterated geometric harmonics method, several points with missing data were selected at random. For each such datapoint, we drew line segments connecting $\ve[x]_i^{(0)}$ (plotted as $\star$) to both $\ve[x]_i^{(end)}$ (plotted as $\bullet$) and $\ve[x]_i$ (plotted as \scalebox{0.6}{$\blacksquare$}). Here, $\ve[x]_i^{(0)}$ means the stochastically initialized version of the point computed before beginning steps (1)--(6), and $\ve[x]_i^{(end)} = \ve[x]_i^{(10)}$ is the final imputed version of the point (the experiment was run for 10 iterations). This provides a graphical indication of how far the algorithm moves some points in order to restore them to their (almost) correct position. In most cases, the imputed point $\ve[x]_i^{(end)}$ is so close to the original point $\ve[x]_i$ that the two are difficult to distinguish. It is clear from the figures that points often required a large correction, and that the algorithm was able to supply this correction.

Since we have the original intact dataset to compare with, we were able to compute the discrepancy between the original and imputed images using dimension-scaled standard $L^2$-norm
\begin{align}\label{eqn:L2-error}
	\text{Error} = p^{-1/2} \sqrt{\sum_{i,j} \left(X(i,j) - \widetilde X(i,j)\right)^2},
\end{align}
where $X$ is the original dataset, $\widetilde X$ is the imputed dataset, and $d$ is the number of dimensions, or parameters, of each data point in the dataset.

The graphs in Figure~\ref{fig:swiss-roll-trails} show the $L^2$ error and standard deviation for values of $p=0.1,0.2,\dots,0.6$. Even with a data annihilation rate of $p=0.50$ (in which case approximately $50\%$ of the data is lost), the $L^2$ error decreases by an order of magnitude in about 4 or 5 iterations, after which it decreases only slightly.

\subsection{Natural data}
\label{sec:natural-data}

\subsubsection{Image datasets}
\label{sec:image-data}
The method was tested on both the Olivetti and UMIST faces datasets, with excellent results. The Olivetti dataset is comprised of 400 photos (10 shots of each of 20 people), each of which is 64$\times$63 pixels. The UMIST dataset is comprised of 15--30 photos of each of 20 people, each of which is 112$\times$92 pixels. Both datasets are freely available on the web.

\begin{figure*}%[h]
	$\negsp[65]$\scalebox{0.375}{\includegraphics{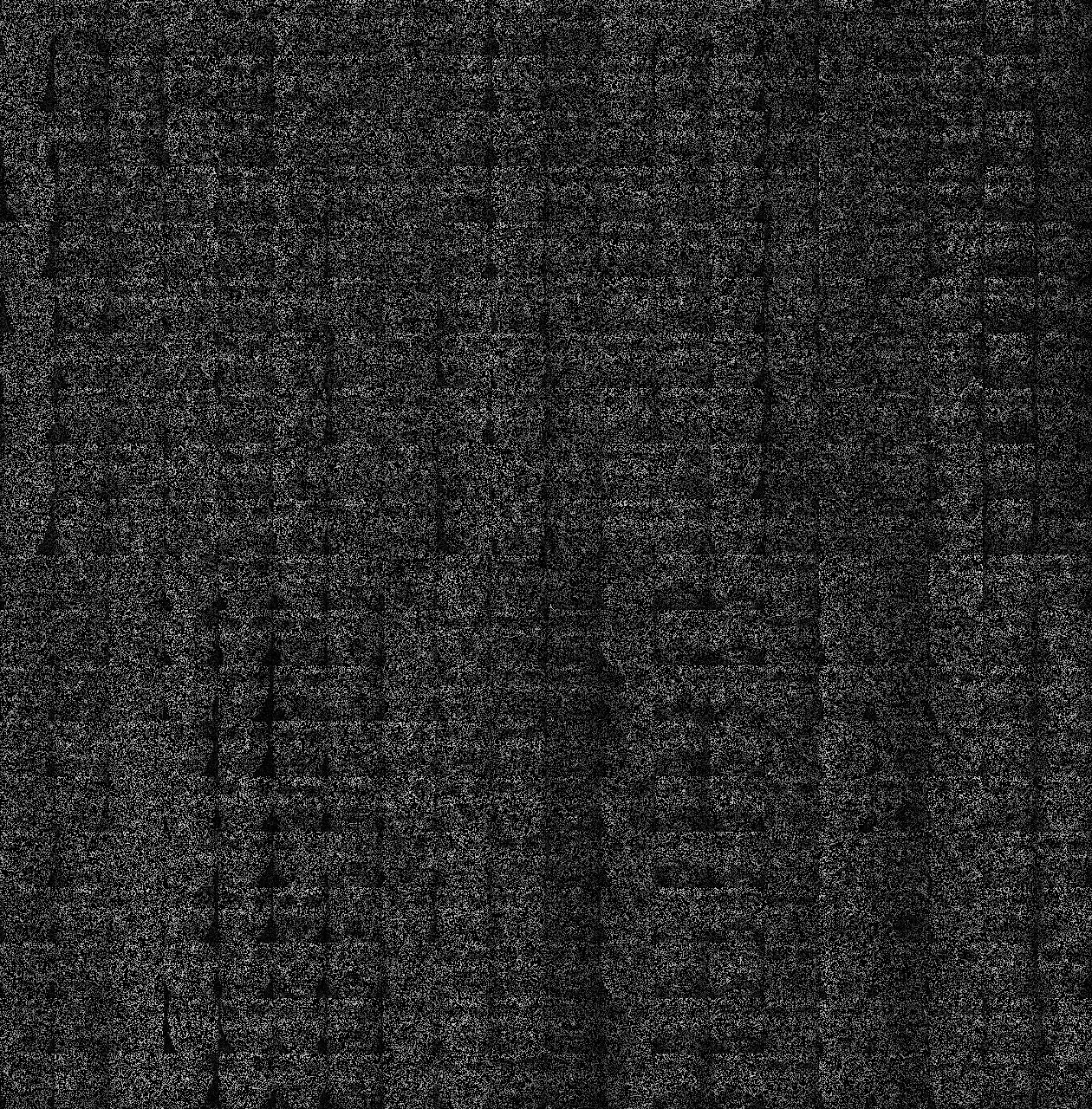}}
	\caption{\captionsize The Olivetti faces dataset with data annihilation rate of $p=0.70$; see \S\ref{sec:natural-data}. Missing data values have been plotted as black pixels, and this is the reason for the apparent poor image quality. This is the data used as input for the method; the resulting output is displayed in Figure~\ref{fig:Olivetti-imputed}.}
	\label{fig:Olivetti-damaged}
\end{figure*}

\begin{figure*}%[h]
	$\negsp[65]$\scalebox{0.375}{\includegraphics{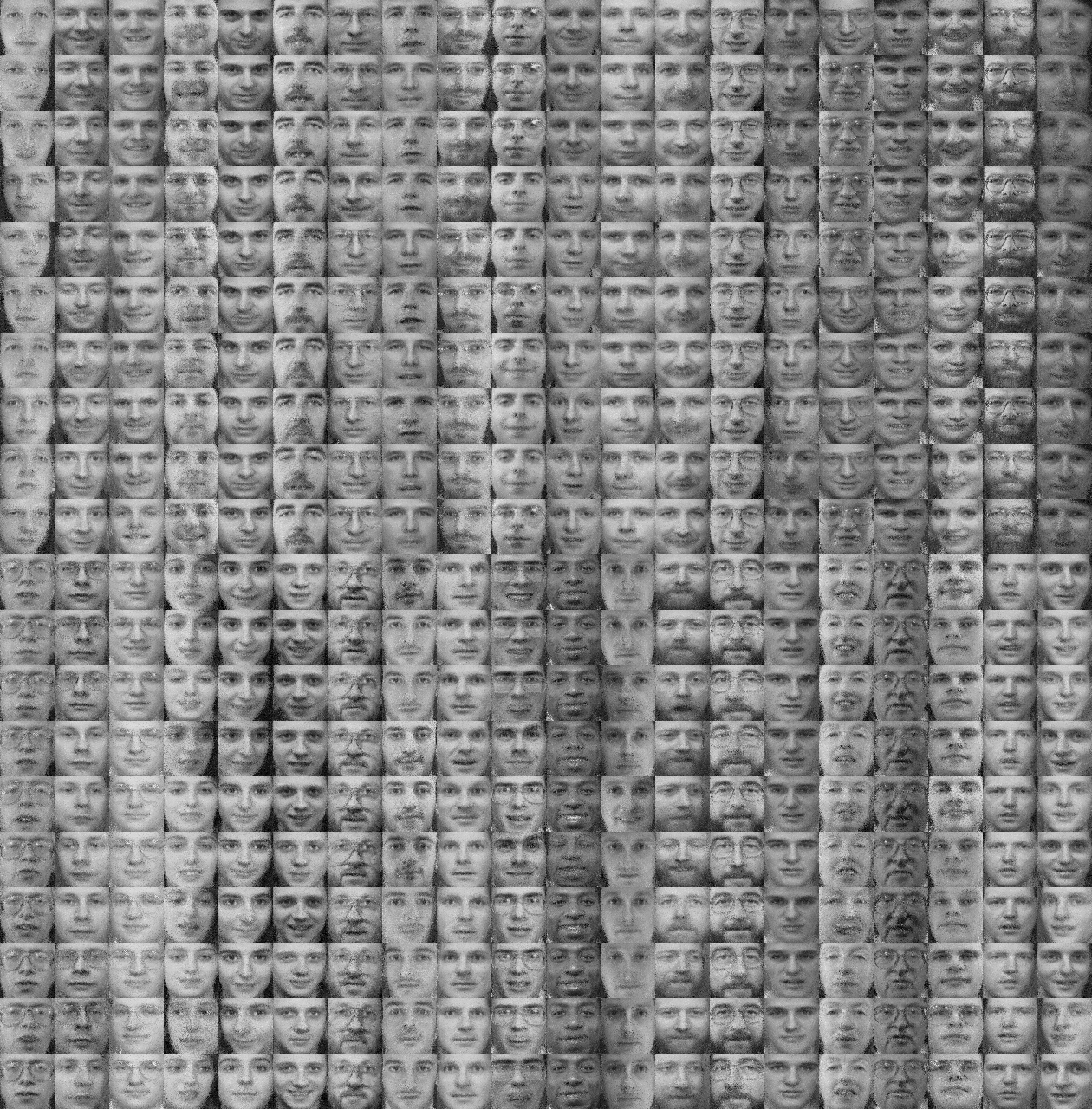}}
	\caption{\captionsize The dataset after imputing missing values using the technique of iterated geometric harmonics; see \S\ref{sec:natural-data}. This is the output that resulted from using the data shown in Figure~\ref{fig:Olivetti-damaged} as input for the method.}
	\label{fig:Olivetti-imputed}
\end{figure*}

To test the method, the test dataset was run through a program which deletes any value in the matrix with probability $p$. For example, Figure~\ref{fig:Olivetti-damaged} shows the result of running this method on the entire Olivetti faces dataset, with data annihilation rate $p=0.70$. Each pixel has probability $p=0.70$ of being deleted, so approximately $70\%$ of the data is lost. Nonetheless, as shown in Figure~\ref{fig:Olivetti-imputed}, IGH is able to reconstruct the data with a startlingly high degree of accuracy: while some noise remains (inevitably), the people in the photos are now clearly recognizable. 

\begin{figure*}%[h]
	$\negsp[65]$\scalebox{0.375}{\includegraphics{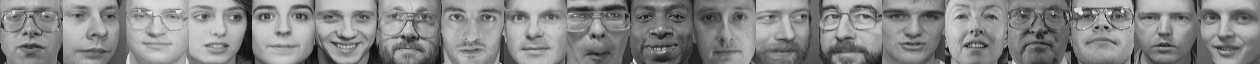}}
	\caption{\captionsize For comparison, this figure shows the original (undamaged) photos from the bottom row of the Olivetti dataset displayed in Figure~\ref{fig:Olivetti-imputed}.}
	\label{fig:Olivetti-original}
\end{figure*}

\begin{figure}%[h]
	\centering
	\scalebox{1.0}{\includegraphics{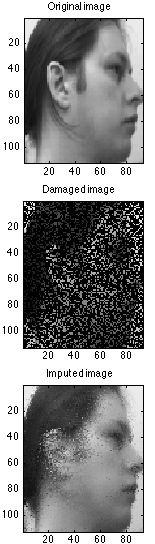}}
	\caption{\captionsize The original, damaged, and imputed (reconstructed) versions of one of the faces in the UMIST dataset. These images are 112$\times$92 pixels and there were 30 images in this cluster (the dataset contains 30 photos of this particular individual). This image highlights two areas where the method has difficulty: (i) divisions between very dark and very light regions, and (ii) areas with larger movement (because of camera angle, the ear moves faster across the frame than nose and lips).}
	\label{fig:UMIST-stages}
\end{figure}

The same procedure was carried out for the UMIST faces database. Figure~\ref{fig:UMIST-stages} shows the original, damaged, and imputed (reconstructed) versions of one of the faces in the dataset.

\begin{remark}\label{def:no-neighbours}
	 The image reconstruction scheme presented here is novel in that a pixel's value is imputed \emph{\textbf{without} making any direct comparison to neighboring pixel values within the same image}. Instead, the pixel's value is inferred based entirely on comparisons with \emph{other} images. In other words, the algorithm makes no use of the fact that the pixel in position $(12, 37)$ in image $i$ is adjacent to pixels $(11,37)$, $(13,37)$, $(12,36)$, or $(12,38)$ of image $i$, and that consequently, these pixels values are likely correlated. %This suggests that it may be possible to improve upon the current results by incorporating additional knowledge of the internal structure of the data points, and the authors are currently pursuing this line of enquiry.
\end{remark}

\begin{figure}%[h]
	\centering
	$\negsp[110]$\begin{minipage}{12cm}
	\scalebox{0.75}{\includegraphics{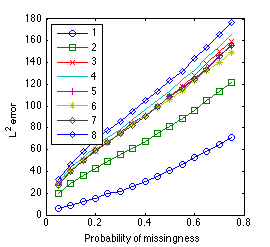}}
	\scalebox{0.75}{\includegraphics{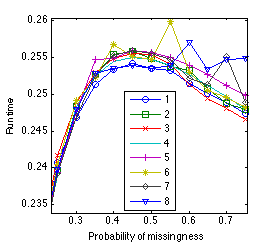}}
	\end{minipage}
	\caption{\captionsize Performance for the COIL-20 dataset. Left: Dependence of accuracy on data sparsity (lower values are better); see \S\ref{sec:image-data}. Right: Dependence of run time on data sparsity (lower values are better); see \S\ref{sec:runtime}. Run time continues to drop sharply as $p$ tends to 0.}
	\label{fig:runtime_vs_density}
	\label{fig:error_vs_density}
\end{figure}

To study the effect of sample density on the performance of the method, we used images from the COIL-20 dataset of $128\times128$ pixel photos of 20 different objects. The full COIL-20 dataset contains 72 photographs of each object, corresponding to increments of $5^\circ$ of rotation. For the experiment reported in Figure~\ref{fig:error_vs_density}, each test was performed using 8 images of the same object: a toy car (object \#3). For ``sparsity 1'', we used a sequence of successive images ($5^\circ$ rotation between samples), for ``sparsity 2'', we used a sequence of every-second images ($10^\circ$ rotation between samples), and so forth. The graph indicates that error increases sharply as sparsity varies from 1 to 3, after which point sparsity makes little difference.

\subsubsection{Weather data}
\label{sec:weather-data}
\begin{figure}%[h]
	\centering
	\scalebox{0.6}{\includegraphics{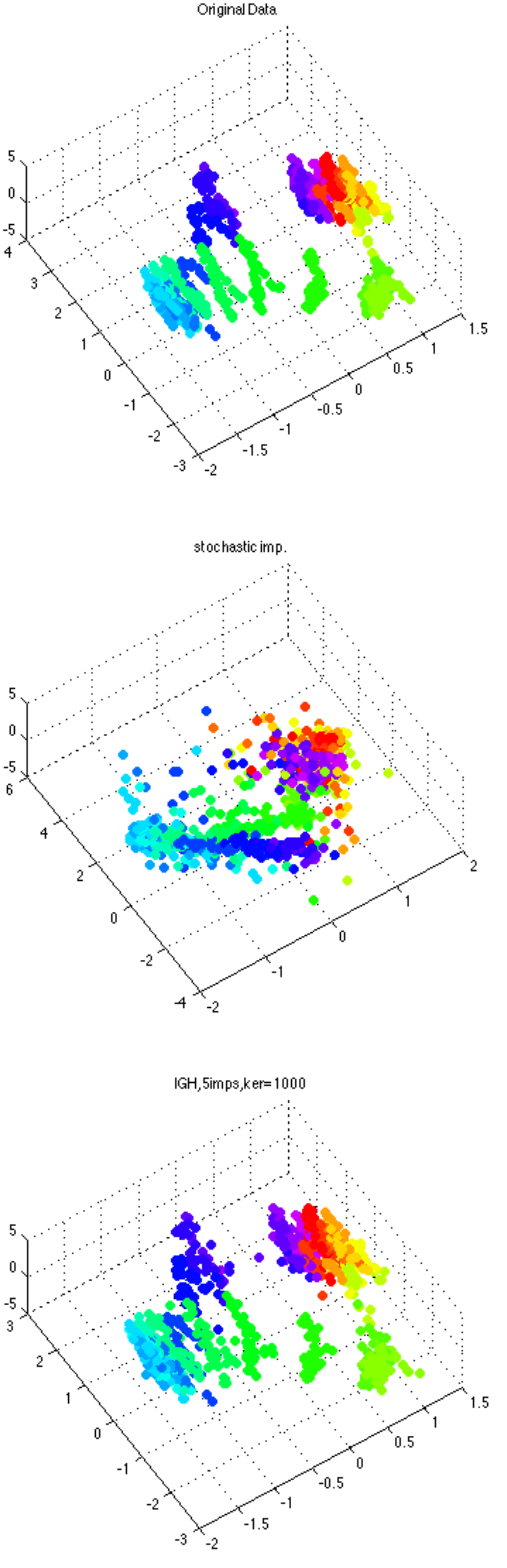}}
	\caption{\captionsize The original, damaged, and imputed (reconstructed) versions of the San Diego weather data.}
	\label{fig:weather}
\end{figure}

\begin{figure}%[h]
	\centering
	\hstr[2]\scalebox{0.45}{\includegraphics{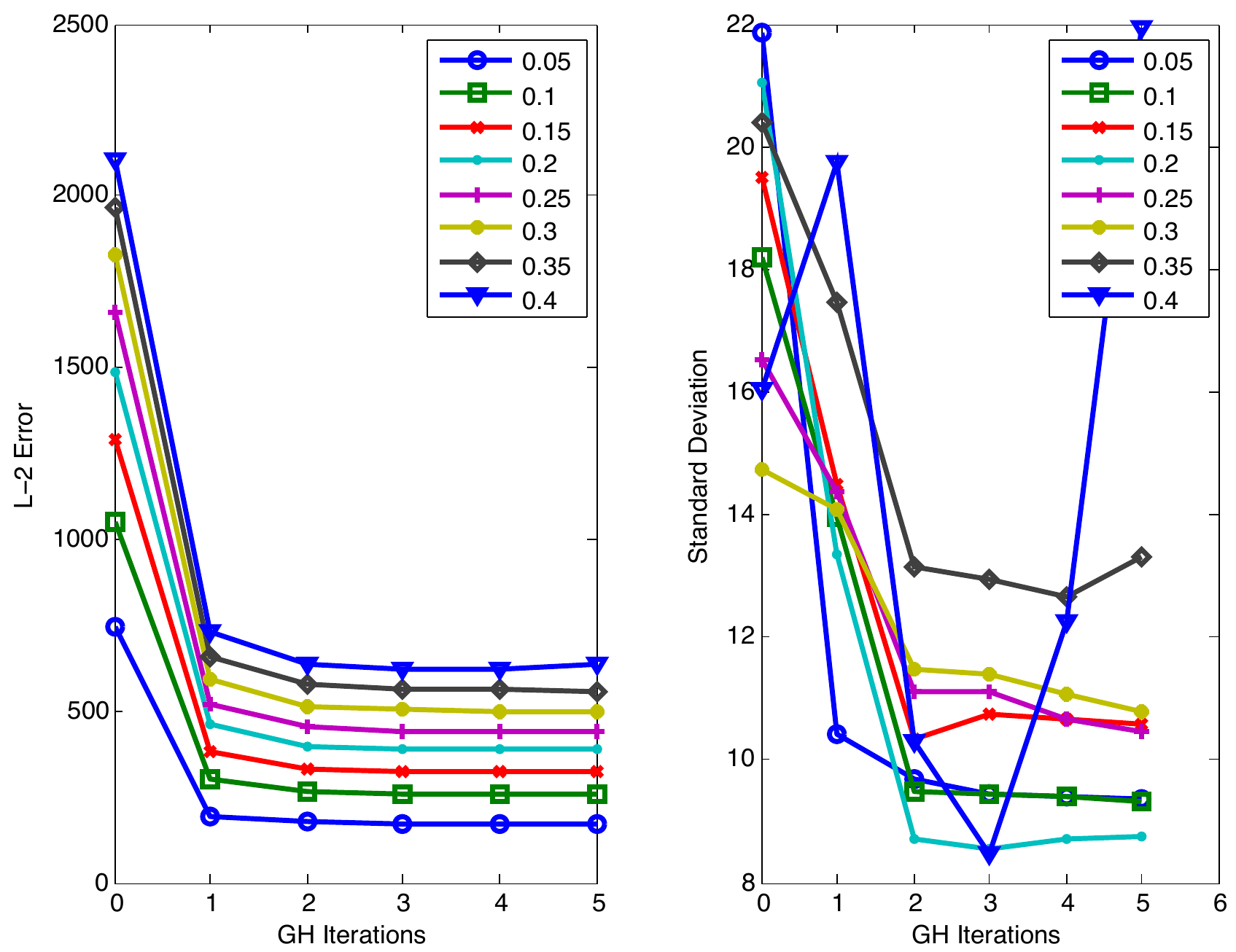}}
	\caption{\captionsize Performance of IGH on the San Diego weather data: the decay of $L^2$ error (left; lower values are better) and standard deviation (right) with each iteration. See \S\ref{sec:weather-data}. }
	\label{fig:weather-performance}
\end{figure}

The algorithm was tested on a set of 2000 data points recorded at San Diego Lindbergh Field's weather station GHCND:USW00023188 (located at the San Diego International Airport: latitude 32.73361, longitude -117.18306) during the period Jan.~1--Mar.~24 of 2010. Each record contains 25 parameters, including average and quantile measurements of temperature, pressure, dew point, wind velocity, and cloud cover, and these measurements were recorded hourly (24 times per day). As with the image datasets, individual measurements were deleted with probability $p$, and the resulting decimated dataset was used as input for the IGH algorithm. Results are presented in Figure~\ref{fig:weather} and Figure~\ref{fig:weather-performance}; in 2--3 iterations, the algorithm reduces the relative error of a random initialization by approximately 70\%. Note that the random initialization is stochastically imputed by drawing from a normal distribution with mean and variance determined by the nonmissing data, exactly as is done during multiple imputation (MI) routines. The standard deviation is not as well-behaved for this experiment, but note that the scaling of the axes magnifies what are actually very small discrepancies.

\subsection{Convergence rates}

The convergence properties of the IGH method are difficult to study analytically, as can be seen from representing the updates in terms of matrix multiplication, as follows. Let the \nth[j] column of $X$ after $t$ iterations be denoted by $\ve[x]_j^{(t)}$. A single iteration of IGH is comprised of doing each of the updates 
\begin{align}\label{eqn:GH-update}
	\ve[x]_j^{(t+1)} \mapsfrom L_j^{(t)} \ve[x]_j^{(t)},
\end{align}
in some random order (see Remark~\ref{rem:shuffle}), where the linear operators $L_j^{(t)}:\bR^n \to \bR^n$ update the \jth column on iteration $t$ according to
\begin{align*}%\label{eqn:}
	L_j^{(t)} = \gY^{(j,t)} (\gy^{(j,t)})^T K_{A_j}^*, \qq j=1,\dots,|A_j|.
\end{align*}
Here, $\gY^{(j,t)}$ is the version of $\gY^{(j)}$ appearing at time $t$, where $\gY^{(j)}$ is the $n \times |A_j|$ matrix whose columns are the geometric harmonics $\gY^{(j)}_\ell$. It is shown in \cite{Lafon04,GeometricHarmonics} that for $A \ci X$, geometric harmonics produces the extension $F:X \to \bR$ of $f:A \to \bR$ which is maximally concentrated on $A$ in the sense that the energy of $F$ on $X \less A$ is minimized. Consequently, each update \eqref{eqn:GH-update} replaces $\ve[x]_j^{(t)}$ with a new version $\ve[x]_j^{(t+1)}$ that has minimal energy on $X \less A_j$. This suggests formulation of a condition for convergence based on a norm which incorporates this energy contribution; every $L_j^{(t)}$ is nonexpansive for  
\begin{align*}%\label{eqn:}
	\| \ve[x]_j\| := \|\ve[x]_j\|_{L^2(X_{A_j})} + \sE_{X \less X_{A_j}}(\ve[x]_j),
\end{align*}
where  $\sE_{X \less X_{A_j}}$ is the restriction of the usual graph energy to the complement of $A_j$.

%However, note that for all $i \neq j$, the entries of $\ve[x]_i^{(t)}$ may differ from those of $\ve[x]_i^{(t-1)}$ due to the previous update. Since the eigendata is constructed in terms of these entries, the operators $L_j^{(t)}$ depend on $t$. As a consequence, the method involves multiplying each $\ve[x]_j$ by a sequence of matrices $\sL_j = (L_j^{(t)})_{t \geq 1}$, and in general this sequence may have infinitely many distinct terms. Thus, we are unable to draw conclusions based on analysis of the joint spectral radius of $\sL_j$. Also, due to the random shuffling (see Remark~\ref{rem:shuffle}), we are unable to make use of strategies similar to those used by Ostrowski-Reich for the method of Successive Over-Relaxations, etc. In light of this, we have only studied convergence properties numerically.

The rate at which the data converges with geometric harmonics decreases approximately linearly with the rate of missing data (i.e., as $p=0.1,0.2,\dots,0.9$). In general, we found that IGH stabilizes after 2 to 10 iterations, with 4 being sufficient in many cases. The bottom left image of Figure~\ref{fig:swiss-roll-trails} shows the average relative $L^2$ error versus number of iterations for swiss rolls ranging from 10 to 60 percent missing values, and the right image in the figure shows the variance of each curve. The general trend is as expected: datasets with more missing values require more iterations to stabilize and are less consistent in their outcome.

\section{Implementation}

MATLAB (R2013a) code for the IGH algorithm, including the examples discussed above, may be found at \\
\phantom{m} \hfill \url{http://www.calpoly.edu/~epearse/IGH.html}. \hfill \phantom{m} \\
%The implementation includes a parameter $\gl_{\text{min}}$ to cope with ill-conditioning (see Remark~\ref{def:Numerical-difficulties}) but we found no significant
Much of this code was written for versatility rather than efficiency, so runtimes could likely be improved. All tests were run on a Mac Pro with 3.2 Ghz Quad-Core Intel Xeon processor, 6GB of 1066 MHz DDR3 RAM, running OS X v10.7.5.

\subsection{Execution speed}
\label{sec:runtime}

\begin{figure}%[h]
	\centering
	\scalebox{0.85}{\includegraphics{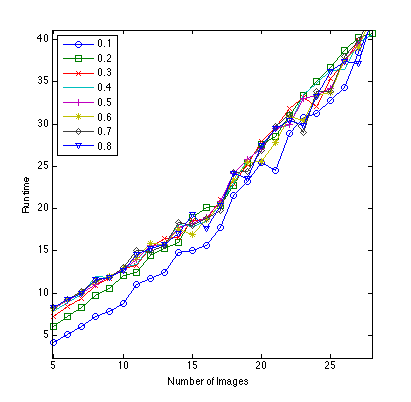}}
	\caption{\captionsize Results of the experiment described in \S\ref{sec:runtime} (lower values are better): run times increase linearly with the number of data points and while slightly shorter for low $p$-values, change very little with $p$.}
	\label{fig:Runtime_averages}
\end{figure}

To test how run time depends on the size of the dataset and on $p$ (the rate of missingness), a number of images were retrieved from the UMIST dataset (so $d=10,304$) and the IGH procedure was applied to this subset, for different values of $p$. Figure~\ref{fig:Runtime_averages} indicates that run time increases approximately linearly with the number of records in the dataset, and that while run time increases with $p$, the effect of missingness is negligible for $p > 0.2$. For each experiment, data annihilation was performed randomly (with the specified probability $p$) and this accounts for the irregularity in the (approximately) linearly increasing graphs.

\bibliographystyle{math}%}plain
\bibliography{network-data}

\newcommand{\etalchar}[1]{$^{#1}$}
\begin{thebibliography}{NLCK2}

\bibitem[Aro]{Aronszajn}
N.~Aronszajn.
\newblock {Theory of reproducing kernels}.
\newblock {\em Trans. Amer. Math. Soc.} {\bf 68}(1950), 337--404.

\bibitem[BN]{Belkin03laplacianeigenmaps}
Mikhail Belkin and Partha Niyogi.
\newblock {Laplacian Eigenmaps for Dimensionality Reduction and Data
  Representation}.
\newblock {\em Neural Computation} {\bf 15}(2003), 1373--1396.

\bibitem[BNS]{BelkinNiyogiSindhwani06}
Mikhail Belkin, Partha Niyogi, and Vikas Sindhwani.
\newblock {Manifold regularization: a geometric framework for learning from
  labeled and unlabeled examples}.
\newblock {\em J. Mach. Learn. Res.} {\bf 7}(2006), 2399--2434.

\bibitem[BPV{\etalchar{+}}]{BengioPaiementVincentDelalleauLeRouxOuimet}
Y.~Bengio, J.F. Paiement, P.~Vincent, O.~Delalleau, N.~Le Roux, and M.~Ouimet.
\newblock {Out-of-sample extensions for LLE, Isomap, MDS, eigenmaps, and
  Spectral Clustering}.
\newblock {\em Advances in Neural Information Processing Systems} {\bf
  16}(2006), 214--225.

\bibitem[CKL{\etalchar{+}}]{MR2443014}
R.~R. Coifman, I.~G. Kevrekidis, S.~Lafon, M.~Maggioni, and B.~Nadler.
\newblock {Diffusion maps, reduction coordinates, and low dimensional
  representation of stochastic systems}.
\newblock {\em Multiscale Model. Simul.} {\bf 7}(2008), 842--864.

\bibitem[CL1]{MR2238665}
Ronald~R. Coifman and St{\'e}phane Lafon.
\newblock {Diffusion maps}.
\newblock {\em Appl. Comput. Harmon. Anal.} {\bf 21}(2006), 5--30.

\bibitem[CL2]{GeometricHarmonics}
Ronald~R. Coifman and St{\'e}phane Lafon.
\newblock {Geometric harmonics: a novel tool for multiscale out-of-sample
  extension of empirical functions}.
\newblock {\em Appl. Comput. Harmon. Anal.} {\bf 21}(2006), 31--52.

\bibitem[DM]{DrineasMahoney2005}
Petros Drineas and Michael~W. Mahoney.
\newblock {On the Nystr\"om Method for Approximating a Gram Matrix for Improved
  Kernel-Based Learning}.
\newblock {\em Journal of Machine Learning Research} {\bf 6}(2005), 2153--2175.

\bibitem[End]{Enders}
Craig~K. Enders.
\newblock {\em Applied Missing Data Analysis}.
\newblock Methodology in the Social Sciences. The Guilford Press, 2010.

\bibitem[FBCM]{FowlkesBelongieChungMalik}
Charless Fowlkes, Serge Belongie, Fan Chung, and Jitendra Malik.
\newblock {Spectral grouping using the Nystr\"om method}.
\newblock {\em IEEE Transactions on Pattern Analysis and Machine Intelligence}
  {\bf 26}(2004), 214--225.

\bibitem[FBM]{FowlkesBelongieMalik}
Charless Fowlkes, Serge Belongie, and Jitendra Malik.
\newblock {Efficient spatiotemporal grouping using the Nystr\"om method}.
\newblock {\em IEEE Comput. Vision Pattern Recogn.} (Dec. 2001).

\bibitem[GM]{GittensMahoney2013}
Alex Gittens and Michael~W. Mahoney.
\newblock {Revisiting the Nystr\"om Method for Improved Large-Scale Machine
  Learning}.
\newblock {\em CoRR} {\bf abs/1303.1849}(2013).

\bibitem[KCLZ]{MR2724431}
Yosi Keller, Ronald~R. Coifman, St{\'e}phane Lafon, and Steven~W. Zucker.
\newblock {Audio-visual group recognition using diffusion maps}.
\newblock {\em IEEE Trans. Signal Process.} {\bf 58}(2010), 403--413.

\bibitem[Laf]{Lafon04}
S.~Lafon.
\newblock {\em Diffusion maps and geometric harmonics}.
\newblock PhD thesis, Yale University, 2004.

\bibitem[LL]{LafonLee06}
S.~Lafon and A.B. Lee.
\newblock {Diffusion maps and coarse-graining: a unified framework for
  dimensionality reduction, graph partitioning, and data set parameterization}.
\newblock {\em Pattern Analysis and Machine Intelligence, IEEE Transactions on}
  {\bf 28}(Sept 2006), 1393--1403.

\bibitem[LR]{LittleRubin02}
Roderick J~A Little and Donald~B Rubin.
\newblock {\em Statistical Analysis with Missing Data}.
\newblock Wiley-Interscience, 2002.

\bibitem[NLCK1]{MR2447229}
Boaz Nadler, Stephane Lafon, Ronald Coifman, and Ioannis~G. Kevrekidis.
\newblock {Diffusion maps---a probabilistic interpretation for spectral
  embedding and clustering algorithms}.
\newblock In {\em Principal manifolds for data visualization and dimension
  reduction}, volume~58 of {\em Lect. Notes Comput. Sci. Eng.}, pages 238--260.
  Springer, Berlin, 2008.

\bibitem[NLCK2]{MR2238669}
Boaz Nadler, St{\'e}phane Lafon, Ronald~R. Coifman, and Ioannis~G. Kevrekidis.
\newblock {Diffusion maps, spectral clustering and reaction coordinates of
  dynamical systems}.
\newblock {\em Appl. Comput. Harmon. Anal.} {\bf 21}(2006), 113--127.

\bibitem[SSSM]{Scholkopf98nonlinearcomponent}
Bernhard Sch\"olkopf, Alexander Smola, Er~Smola, and Klaus-Robert M\"uller.
\newblock {Nonlinear Component Analysis as a Kernel Eigenvalue Problem}.
\newblock {\em Neural Computation} {\bf 10}(1998), 1299--1319.

\bibitem[SP]{MR0140732}
D.~Slepian and H.~O. Pollak.
\newblock {Prolate spheroidal wave functions, {F}ourier analysis and
  uncertainty. {I}}.
\newblock {\em Bell System Tech. J.} {\bf 40}(1961), 43--63.

\bibitem[Sle]{MR0181766}
David Slepian.
\newblock {Prolate spheroidal wave functions, {F}ourier analysis and
  uncertainity. {IV}. {E}xtensions to many dimensions; generalized prolate
  spheroidal functions}.
\newblock {\em Bell System Tech. J.} {\bf 43}(1964), 3009--3057.

\bibitem[vB]{vanBuuren}
Stef van Buuren.
\newblock {\em Applied Missing Data Analysis}.
\newblock Chapman \& Hall/CRC Interdisciplinary Statistics. Chapman and Hall,
  2012.

\bibitem[WS]{WilliamsSeeger01}
C.~Williams and M.~Seeger.
\newblock {Using the Nystr\"om Method to speed up kernel machines}.
\newblock {\em Neural Inf. Process. Systems} {\bf 13}(2001), 682--688.

\end{thebibliography}

%\newpage

%\begin{IEEEbiographynophoto}{Chad~Eckman}
%received the BSc degree in Mathematics from California Polytechnic State University, San Luis Obispo, where he is currently a graduate student in Mathematics.
%\end{IEEEbiographynophoto}
%
%\begin{IEEEbiographynophoto}{Jonathan~A.~Lindgren}
%received the BSc degree in Engineering from California Polytechnic State University, San Luis Obispo, where he is now a graduate student in Mathematics.
%\end{IEEEbiographynophoto}
%
%\begin{IEEEbiographynophoto}{Erin~P.~J.~Pearse}
%received the PhD degree in Mathematics from University of California, Riverside, California in 2006, and has been with the Department of Mathematics at California Polytechnic State University, San Luis Obispo, since 2012. His research interests include: analysis and geometry of fractals, especially tube formulas and measurability issues; infinite graphs, especially topics related to effective resistance, graph energy and Laplacians, and random walks; and graph-based models of data analysis, especially applications to dimensionality reduction and the imputation of missing data.
%\end{IEEEbiographynophoto}
%
%\begin{IEEEbiographynophoto}{David~J.~Sacco}
%received the BSc degree in Mathematics from California Polytechnic State University, San Luis Obispo, and is currently a graduate student at Oklahoma State University in Mathematics.
%\end{IEEEbiographynophoto}
%
%\begin{IEEEbiographynophoto}{Zachariah~Zhang}
%is currently an undergraduate in Mathematics and Computer Science at California Polytechnic State University, San Luis Obispo.
%\end{IEEEbiographynophoto}
%
%\vfill

\end{document}